\documentclass[final]{cvpr}

\usepackage{times}
\usepackage{epsfig}
\usepackage{graphicx}
\usepackage{amsmath}
\usepackage{amssymb}

\usepackage[dvipsnames]{xcolor}
\usepackage[ruled,vlined]{algorithm2e}
\usepackage{amsmath}
\usepackage{booktabs}
\usepackage{multirow}
\usepackage{flushend}

\usepackage[pagebackref=true,breaklinks=true,colorlinks,bookmarks=false]{hyperref}

\def\cvprPaperID{5562} 
\def\confYear{CVPR 2021}
\newcommand{\imagerep}{\mathcal{I}}

\newcommand{\objectlabelrepeq}{\mathbf{O}}
\newcommand{\relationlabelrepeq}{\mathbf{R}}

\newcommand{\noderep}{$\mathbf{n}_{i}$}
\newcommand{\edgerep}{$\mathbf{e}_{i \rightarrow j}$}
\newcommand{\imagegraph}{$\mathsf{G}_{\imagerep}$}
\newcommand{\imagegrapheq}{\mathsf{G}_{\imagerep}}
\newcommand{\scenegraph}{$\mathsf{G}_{SG}$}
\newcommand{\scenegrapheq}{\mathsf{G}_{SG}}

\begin{document}

\title{Energy-Based Learning for Scene Graph Generation}

\author{
Mohammed Suhail$^{1,2}$ \qquad Abhay Mittal$^{4}$ \qquad Behjat Siddiquie$^{4}$  \qquad Chris Broaddus$^{4}$ \\ Jayan Eledath$^{4}$  \qquad Gerard Medioni$^{4}$ \qquad Leonid Sigal$^{1,2,3}$\\
$^1$University of British Columbia \qquad
$^2$Vector Institute for AI \qquad
$^3$Canada CIFAR AI Chair \qquad
$^4$ Amazon \\
{\tt\small suhail33@cs.ubc.ca} \qquad {\tt\small mrmittal@amazon.com} \qquad {\tt\small behjats@amazon.com} \qquad {\tt\small chrispb@amazon.com} \\ \qquad {\tt\small eledathj@amazon.com} \qquad {\tt\small medioni@amazon.com} \qquad {\tt\small lsigal@cs.ubc.ca}
}

\maketitle
\begin{abstract}
Traditional scene graph generation methods are trained using cross-entropy losses that treat objects and relationships as independent entities. Such a formulation, however, ignores the structure in the output space, in an inherently structured prediction problem.
In this work, we introduce a novel energy-based learning framework for generating scene graphs. The proposed formulation allows for efficiently incorporating the structure of scene graphs in the output space. This additional constraint in the learning framework acts as an inductive bias and allows models to learn efficiently from a small number of labels. 
We use the proposed energy-based framework \footnote{Code and pre-trained models available at \url{https://github.com/mods333/energy-based-scene-graph}.} to train existing state-of-the-art models and obtain a significant performance improvement, of up to 21\% and 27\%, on the Visual Genome \cite{krishna2017visual} and GQA \cite{hudson2018gqa} benchmark datasets, respectively. Furthermore, we showcase the learning efficiency of the proposed framework by demonstrating superior performance in the zero- and few-shot settings where data is scarce.
\end{abstract}

\section{Introduction}
\label{sec:introduction}
\begin{figure}[t]
    \centering
    \includegraphics[width=.9\linewidth]{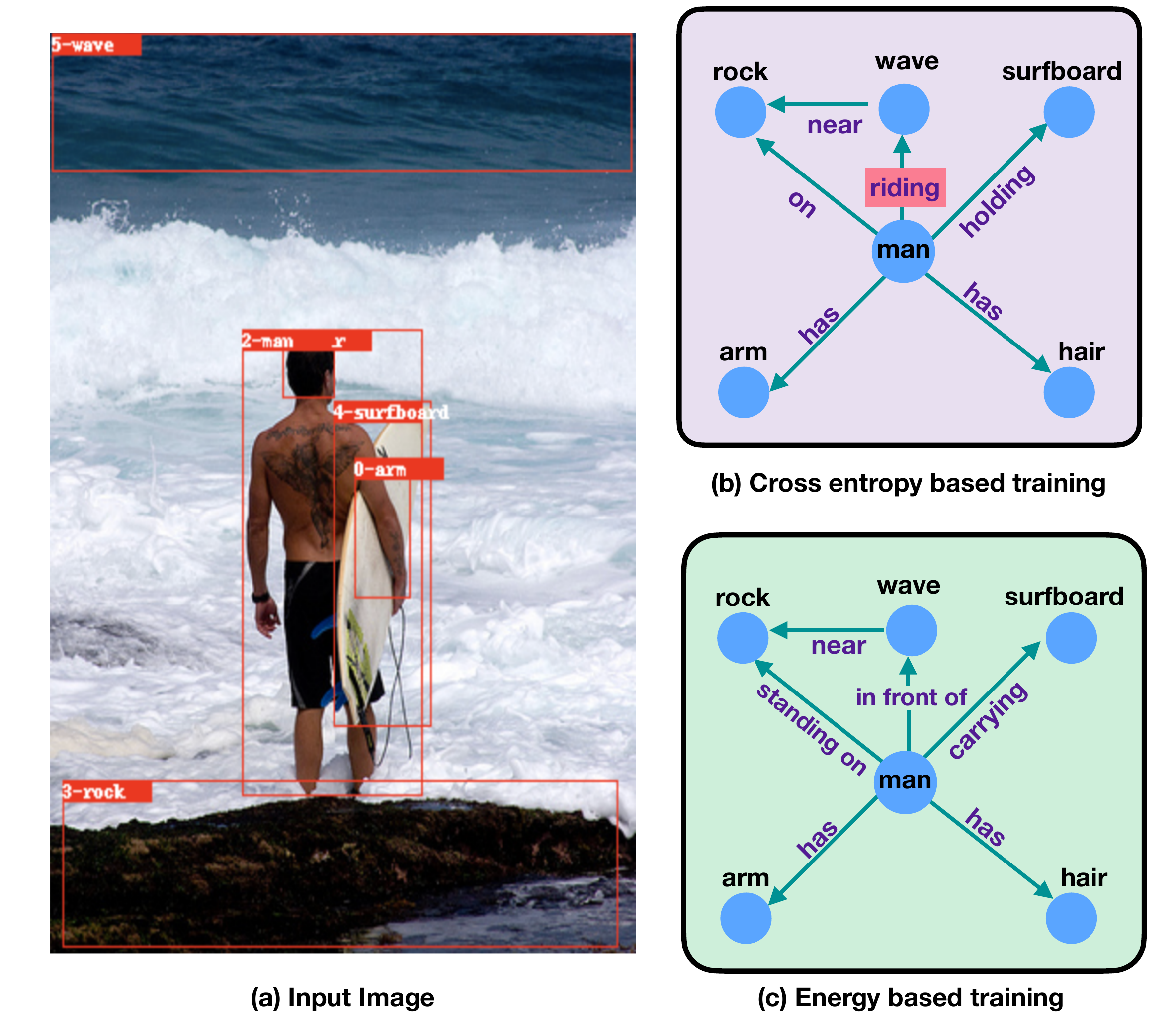}
    \caption{\textbf{Scene Graph Generation}: Figure shows scene graphs generated by a VCTree \cite{tang2019learning} model trained using conventional cross-entropy loss (purple) and our proposed energy-based framework (green). We make two crucial observations. First, the model trained using cross-entropy loss is incapable of consistent structural reasoning ({\tt \small riding} is not possible given the rest of the graph). Second, the trained model tends to be biased, favoring more frequent relations (\eg, {\tt \small on}).
    Our proposed energy-based framework is designed, and able, to address these shortcomings.}
    \label{fig:intro_figure}
\end{figure}

A scene graph is a graph-based representation of an image which encodes objects along with the relationships between them. Such a representation allows for a comprehensive understanding of images that is useful in several vision applications, including visual question answering \cite{hudson2018gqa, tang2019learning}, image captioning \cite{gu2019unpaired, yang2019auto} and scene synthesis \cite{herzig2019learning, johnson2018image}.

A typical scene graph generation model comprises of the object detection network, which extracts object regions and corresponding features, and a message passing network with nodes initialized with these region features and edges accounting for the potential relations among them. 
The features are refined, through context aggregation, and then classified to produce both object (node) and relation (edge) labels. These networks are often trained end-to-end by minimizing individual cross-entropy losses on both sets of labels. 
A major drawback of such an approach is that quality of prediction (loss) is simply proportional to the number of correctly predicted labels and ignores the rich structure of the scene graph output space (\eg, correlation or exclusion among object and relation label sets). 
In addition, the imbalance in the number of training samples for the relations results in dominant relations being heavily favored, leading to biased relation prediction at test time \cite{tang2020unbiased}.

Figure \ref{fig:intro_figure} (b) illustrates the scene graph generated by a model \cite{tang2019learning} trained using the cross-entropy loss. Both the aforementioned drawbacks are apparent in the output. 
First, the model predicts a relation \texttt{<man, riding, wave>}. A simple examination of the rest of the scene graph reveals that such a relationship is impossible given that the \texttt{man} is \texttt{on} a \texttt{rock} and \texttt{holding} a \texttt{surfboard}. Second, the model leans towards making generic relation predictions such as \texttt{<man, on, rock>} as opposed to more informative alternatives, \eg, \texttt{<man, standing on, rock>}. 

The origin of these issues can be identified by examining the likelihood term. Cross-entropy based training treats objects ($O$) and relationships ($R$) in a scene graph as independent entities. This amounts to factorizing the likelihood of a scene graph ($SG$), given an image ($I$), as the product of the likelihoods for the individual objects and relations:
\begin{equation}
    \log p(SG|I) = \sum_{i \in O} \log p(o_i| I) + \sum_{j \in R} \log p(r_j | I).
    \label{eq:ce_likelihood}
\end{equation}
Eq.(\ref{eq:ce_likelihood}) brings to light the underlying cause of the problem highlighted above. First, during loss computation, the loss for each relation term is independent of the relations predicted in the rest of the scene graph.  Thus an incorrect relation such as \texttt{<man, riding, wave>} is penalized the same as \texttt{<man, behind, wave>} irrespective of the other relations (\texttt{<man, on, rock>}). However, using common sense reasoning, we can determine that \texttt{<man, riding, wave>} is highly improbable given \texttt{<man, carrying, surfboard>} and should be penalized heavily as opposed to a likely, albeit incorrect, relation \texttt{behind}. Second, due to the summation over individual relation terms, the model, in order to minimize the loss, is incentivized to predict relations which are more common in the training data.


While prior works have tried to address the issue of biased predictions \cite{Lin_2020_CVPR, tang2020unbiased} in the context of scene graph generation, little progress has been made towards structured learning of scene graphs. 
In this work, we address both of these issues by proposing a novel generic loss formulation that incorporates the structure of scene graphs into the learning framework using an energy-based learning framework. This energy-based framework relies on graph message passing algorithm for energy computation, that is learned to model the joint conditional density of a scene graph, given an image.
Such a formulation transforms the problem from maximizing sum of the individual likelihood terms to that of directly maximizing the joint likelihood of the objects and relations.
Furthermore, this added structure acts as an inductive bias for the learning, allowing the model to efficiently learn relationship statistics from less data. 

The proposed learning framework is general 
and hence can be used to train any off-the-shelf scene graph generation model. We experiment with various state-of-the-art models and demonstrate that our energy-based formulation achieves significant improvements in the performance over the corresponding models trained using the standard cross-entropy based formulation. We also demonstrate the enhanced generalization capability of models trained using our framework by evaluating zero shot relation retrieval performance. Finally, we demonstrate the ability of our energy-based framework to learn from lesser amounts of training data by demonstrating an improvement in relative performance when evaluating on few-shot relation triplets. 

Figure \ref{fig:intro_figure} (c), shows the scene graph generated by the proposed method. The generated scene graph is more granular, predicting relations such as \texttt{<man, standing on, rock>} as opposed to the biased and generic variant \texttt{<man, on, rock>}. The model is also able to preclude improbable relations (\eg, \texttt{<man, riding, wave>}) and instead predicts \texttt{in front of} between \texttt{man} and \texttt{wave}.

\vspace{0.06in}
\noindent
\textbf{Contribution:} Our main contribution is a novel energy-based framework for scene graph generation that allows for direct incorporation of structure into the learning. We also propose a novel message passing algorithm that is used for computing the energy of scene graph configurations. This message passing algorithm is generic and can be used for other applications such as learning graph embeddings. Finally, we demonstrate the efficacy of our proposed framework by applying it to multiple state-of-the-art models and evaluating performance on 
two benchmark datasets - Visual Genome \cite{krishna2017visual} and GQA \cite{hudson2018gqa} - where we consistently outperform the cross-entropy based counterparts by up to $21\%$ on Visual Genome and $27\%$ on GQA.

\section{Related Work}
\noindent
\textbf{Scene Graph Generation:}
Scene Graph generation has become increasingly popular in the vision community  \cite{knyazev2020graph, li2018fnet, Lin_2020_CVPR, newell2017pixels, tang2020unbiased, xu2017scene, yang2018graph, ZareianKC20, ZareianWYC20, zellers2018neural}. Early works, \eg,  \cite{xu2017scene, yang2018graph, zellers2018neural}, focused on improving context aggregation modules that facilitated learning better representations, thereby improving performance. Recent works, \cite{knyazev2020graph, Lin_2020_CVPR, zhang2019graphical}, identify the shortcomings of training using cross-entropy loss and propose handcrafted loss formulations to improve the performance. Our work can be considered as a generalization of the latter, where we allow the model to learn an approximate joint distribution over the scene graphs and images using an energy based formulation. Furthermore, to the best of our knowledge, our work is the first to incorporate structure in the output space, into the learning paradigm.

\vspace{0.08in}
\noindent
\textbf{Energy Based Modeling:}
Energy Based Models(EBM) learn the unnormalized density of the data space, thereby avoiding the need to compute the partition function \cite{lecun2006tutorial}. EBMs have recently sparked interest in generative modeling. One line of work uses energy models to learn the underlying data distribution and then implicitly generates samples from the learned distribution \cite{du2019implicit, grathwohl2020your}. In \cite{song2019generative}, they propose denoising score matching based training, where the model learns to generate data by denoising progressively increasing noisy signals. While there has been an increase in the use of energy-based modeling for generative tasks, they have been relatively unexplored for discriminative tasks. In our work, we show how one can formulate a discriminative task of scene graph generation using an EBM framework.

\section{Approach}
We first provide an overview of current approaches for scene graph generation based on the standard cross entropy loss, followed by a  description of our proposed energy-based learning framework along with the architecture used for energy computation.
\begin{figure*}[t]
    \centering
    \includegraphics[width=.95\linewidth]{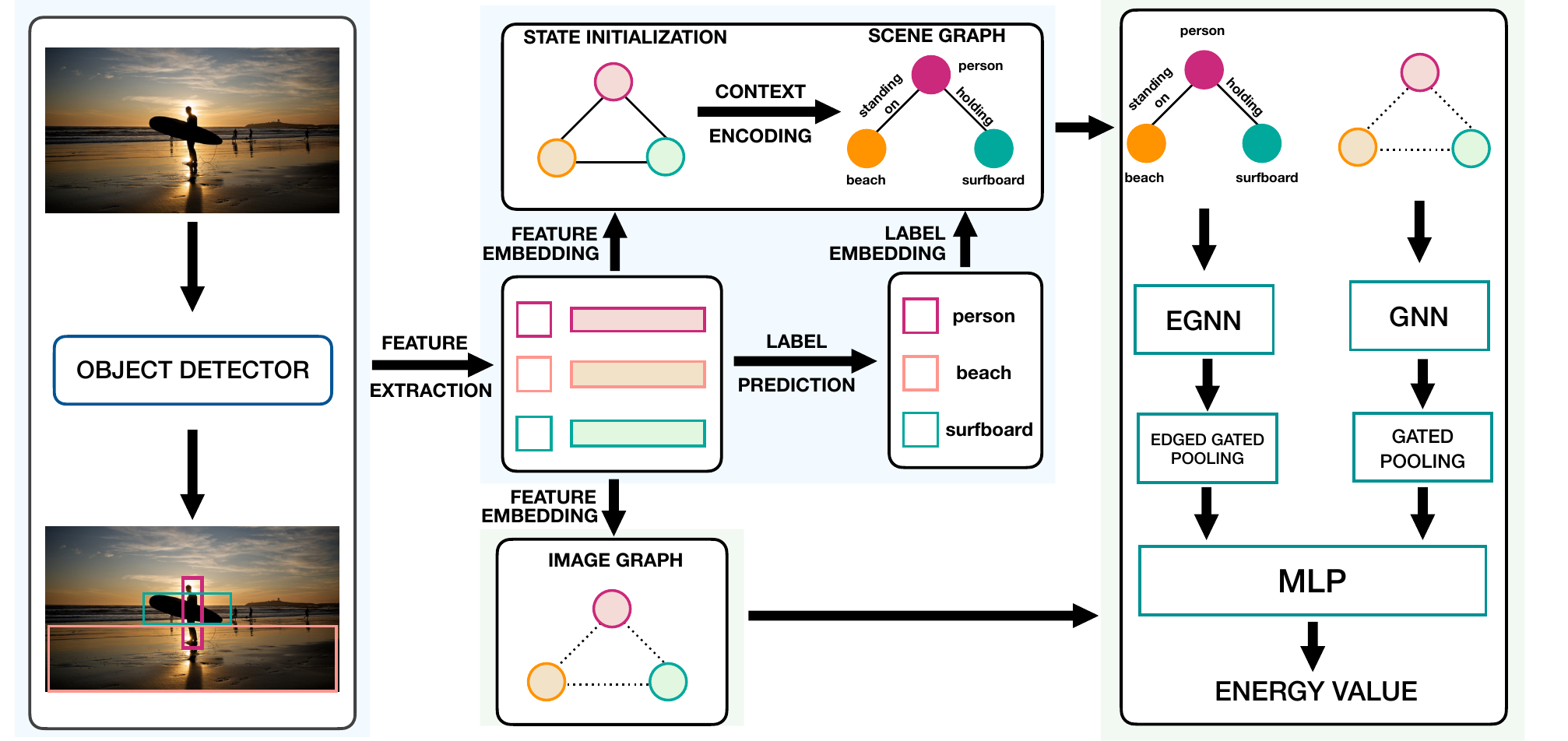}
    \caption{\textbf{Model Overview of the Energy-based Learning.} 
    The region in light blue correspond to most traditional scene graph generation pipelines. The proposed energy-based learning framework is highlighted in light green. We initialize the image graph with the extracted object proposal features as the node states. We instantiate the scene graph with
    predictions from traditional pipeline (or ground truth annotation). The image graph and scene graph are fed into the energy model where they undergo state refinement using a Gated Graph Neural Network and a novel Edged Graph Neural Network, respectively. We then obtain vector representations of each graph using pooling layers. The representations are concatenated and passed as input to a multi-layer perceptron which predicts the energy of the joint input (image) - output (scene graph) configuration. The loss is computed from the energy values of the ground truth and predicted configuration. 
    }
    \label{fig:model_overview}
\end{figure*}
\subsection{Scene Graph Generation}
Scene graph generation methods typically adopt a two stage framework. In the first stage, given an image $\imagerep$, bounding boxes 
are obtained using a standard object detector such as Faster R-CNN \cite{ren2015faster}. Features
corresponding to these regions are extracted using {\tt RoIAlign} along with an initial distribution over object labels. 
In the next stage, these detections are used as inputs to predict the scene graphs. The bounding-box features along with the object label and the spatial coordinates of the bounding boxes are used to initialize a set of node features. These features are refined using architectures such as LSTM \cite{zellers2018neural}, TreeLSTM \cite{tang2019learning} or Graph Attention Networks \cite{yang2018graph}, to incorporate contextual information. Object labels, $\objectlabelrepeq$, are then obtained by classifying the refined features. Relationship labels, $\relationlabelrepeq$, are obtained by extracting features from union of object bounding boxes, followed by state refinement using BiLSTMs \cite{zellers2018neural} or BiTreeLSTMs \cite{tang2019learning} and subsequent classification. 

These models are trained using standard cross-entropy loss on object and relation labels. Each object and relationship is considered in isolation when computing individual losses, which are then summed up to obtain the loss for the given image. Such a loss formulation ignores the fact that objects and relations in a scene graph are inter-dependent. Intuitively, incorporating such dependencies into the learning procedure should lead to an improvement in performance. However, it is not clear as to how one can exploit the rich structure in the output space. Most methods (\cite{tang2019learning,xu2017scene, yang2018graph,zellers2018neural}) attempt to find a way around this by employing message passing algorithms in the input space that allow for aggregation of context information. However, this does not explicitly consider structure in the output space; neither in predictions, nor in the loss function used for learning. 
In this work, we propose a novel energy-based learning framework that allows scene graph generation models to be trained using a ``loss" that explicitly incorporates structure in the output space.

\subsection{Energy Based Modeling}
Energy-based models \cite{lecun2006tutorial} encode dependencies between variables by assigning a scalar energy value to an input configuration. Given a data point $\mathbf{x} \in \mathcal{X}$ with corresponding label $\mathbf{y} \in \mathcal{Y}$, let $E_{\theta}(\mathbf{x},\mathbf{y}) \in \mathbb{R}$ be a joint energy function. While energy models map inputs to unnormalized densities, we can define a probability distribution via the Boltzmann distribution, $ p_{\theta}(\mathbf{x},\mathbf{y}) = \frac{\exp(-E_{\theta}(\mathbf{x},\mathbf{y}))}{Z(\theta)}$, where $Z(\theta) = \int \exp(-E_{\theta}(\mathbf{x},\mathbf{y}))$ is referred to as the normalization constant or partition function. Computing the normalization constant, $Z_{\theta}$ for most parameterizations of the energy function is intractable. Therefore, learning the parameters $\theta$ using methods such as maximum likelihood is not straightforward. Most methods address this problem by rewriting the derivative of the log likelihood as $\nabla_{\theta} \log p_{\theta}(\mathbf{x}, \mathbf{y}) = \mathbb{E}_{p_{\theta}(\mathbf{x'},\mathbf{y'})} [ \nabla_{\theta} E_{\theta} (\mathbf{x}, \mathbf{y'})] - \nabla_{\theta} E_{\theta}(\mathbf{x}, \mathbf{y}),$
where the expectation is approximately estimated using MCMC methods that sample from the data distribution. 

Unlike most prior works that train energy models for generative modeling, our focus is scene graph generation, a discriminative task. For such a task, we are only concerned with the relative energies of the various label configurations given an input $\mathbf{x}$. Training with a carefully crafted loss function circumvents the need for estimating the partition function or computing expectations. Therefore we can parametrize the energy function using an arbitrary neural network architecture. For a more detailed discussion on energy loss formulation refer to Section 2 in \cite{lecun2005loss}. 

\subsection{Energy Models for Scene Graphs Generation}
We now describe our proposed energy-based learning framework for scene graph generation. In our formulation of the energy function, the data space $\mathcal{X}$ is the set of images $\imagerep \in \mathbb{R}^{W \times H \times 3}$ and the label space $\mathcal{Y}$ is the set of scene graphs $SG$. 
The scene graph, $SG$, is defined by a tuple  $(\objectlabelrepeq,\relationlabelrepeq)$, where $\objectlabelrepeq \in \mathbb{R}^{n \times d}$ is the set of object labels and $\relationlabelrepeq \in \mathbb{R}^{n \times n \times d'}$ is the set of relationship labels; $n$ is the number of objects in an image; $d$ and $d'$ is the total number of possible object and relation labels in the dataset. 


A simple implementation of the joint energy function, would take an encoding of the image, $\imagerep$, and a scene graph, $SG$, and produce an scalar energy value. However, there are a few challenges with this. First, a simple global CNN-based encoding of the image may fail to highlight, potentially small, regions relevant for scene graph prediction. Second, scene graph representation is variable in length ($n$ is not fixed) and high dimensional. The second challenge can be addressed by pooling object $\objectlabelrepeq$ and relations $\relationlabelrepeq$ across $n$ and $n \times n$ dimensions respectively. We propose a more sophisticated and effective scene graph refinement and gated pooling formulation in  Section~\ref{sec:energy_model_architecture}. 
To facilitate the former challenge of image encoding, we extract a graph based representation from the image.
This representation is henceforth referred to as an image graph (\imagegraph).
The nodes of the image graph are instantiated using features extracted from the object bounding boxes.


Given a scene graph generation model $M$ and an image $\imagerep$, we predict a scene graph, $\scenegrapheq^{0}$ \  and compute the image graph \imagegraph. The scene graph along with the image is provided as input to the energy model ($E_{\theta}$) to compute the energy corresponding to the predicted configuration. Similarly, we compute the energy of the ground truth configuration using the ground truth scene graph  ($\scenegrapheq^{+}$) and image graph ($\imagegrapheq^{+}$) constructed from ground truth bounding boxes.  These two energy values are then used to compute the energy loss
\begin{equation}
    \mathcal{L}_{e} = E_{\theta}(\imagegrapheq^{+}, \scenegrapheq^{+}) - \min_{\scenegrapheq \in SG} E_{\theta}(\imagegrapheq, \scenegrapheq). 
    \label{eq:energy_loss}
\end{equation}
Computing this loss requires solving an optimization problem to find a scene graph configuration that minimizes the energy value (second term in Eq.(\ref{eq:energy_loss})). Starting from $\scenegrapheq^{0}$ we use Stochastic Gradient Langevine Dynamics (SGLD) \cite{welling2011bayesian} which approximately solves the optimization problem in an iterative manner: 
\begin{align}
\objectlabelrepeq^{\tau+1} &= \objectlabelrepeq^{\tau} - \frac{\lambda}{2} \nabla_{\objectlabelrepeq}E_{\theta}(\imagegrapheq, \scenegrapheq^{\tau}) + \epsilon^{\tau},  \nonumber \\
\relationlabelrepeq^{\tau+1} &= \relationlabelrepeq^{\tau} - \frac{\lambda}{2} \nabla_{\relationlabelrepeq}E_{\theta}(\imagegrapheq, \scenegrapheq^{\tau}) + \epsilon^{\tau}
\label{eq:sgld}
\end{align}
where $\objectlabelrepeq^{t}$ and $\relationlabelrepeq^{t}$ are the node and edge states in the scene graph and $\epsilon^{t}$ is sampled from a normal distribution $\mathcal{N}(0,\lambda)$. 

Conceptually, the predicted $\scenegrapheq^{0}$ is used as an ``initialization" to arrive at the low energy configuration through a series of steps defined by Eq.(\ref{eq:sgld}); each step similar to regular gradient descent with an added Gaussian noise. Differentiation through the optimization path guides parameter learning of the model $M$ that generates $\scenegrapheq^{0}$ in the first place. 


When training using the above loss, we observe that the energy values get arbitrarily large in magnitude leading to gradient overflow.  We address this problem by adding an $L2$ regularization loss on the energy values:
\begin{equation}
    \mathcal{L}_{r} = E_{\theta}(\imagegrapheq^{+}, \scenegrapheq^{+})^{2} + E_{\theta}(\imagegrapheq, \scenegrapheq)^{2}.
\end{equation}

Finally, since the space of scene graphs is very high dimensional, we need to restrict the search space of the energy model in order to stabilize the learning. This is done by incorporating the task loss used by the underlying scene graph generation model, $\mathcal{L}_{t}$ on the predicted output as an added regularization on the initial prediction. The total loss for training the scene graph generator and the the energy model is given by:
\begin{equation}
    \mathcal{L}_{total} = \lambda_{e}\mathcal{L}_{e} + \lambda_{r}\mathcal{L}_{r} + \lambda_{t}\mathcal{L}_{t},
\end{equation}
where $\lambda_{e}$, $\lambda_{r}$ and $\lambda_{t}$ are the relative weights.  

\subsection{Energy Model Architecture}
\label{sec:energy_model_architecture}
Given an image graph (\imagegraph) and a scene graph (\scenegraph), the energy
model first refines the state representations using graph neural networks. We use a novel Edged Graph Neural Networks (EGNN) and Graph Neural Network \cite{DBLP:journals/corr/LiTBZ15} on the scene graph and image graph respectively to incorporate contextual information. This is followed by applying a pooling layer on each graph to obtain a vector representation summarizing the graph states. Finally, these two vectors are fed into a multi layer perceptron (MLP) to compute the energy value corresponding to the predicted scene graph configuration. We then repeat these operations for the ground truth scene graph and the input image. The energy model can be parametrized as:
\vspace{-0.05in}
\begin{align}
    E_{\theta}(\imagegrapheq, \scenegrapheq) &= \mathsf{MLP} \left[ f(\mathtt{EGNN}(\scenegrapheq)); g(\mathtt{GNN}(\imagegrapheq)) \right],
\end{align}
where $f$ and $g$ are pooling functions.

\vspace{0.08in}
\noindent
\textbf{Notation.} 
We use \noderep~to represent the features of the $i^{th}$ node, which for $\scenegrapheq$ corresponds to the $i^{th}$ object, initialized to the  corresponding $i^{th}$ row of matrix $\objectlabelrepeq$. For $\imagegrapheq$  \noderep~corresponds to the $i^{th}$ image region, initialized by the {\tt RoIAlign} image features. 
We use \edgerep~to represent the feature of the directed edge from node $i$ to $j$, initialized to $(i,j)^{th}$ column of $\relationlabelrepeq$. $\mathcal{N}_{i}$ denotes the neighbours of node ${i}$. 
\subsubsection{Edge Graph Neural Network}
To allow for direct application of convolution operations on graph representations accommodating edge features such as scene graphs, we propose a variant of graph message passing algorithm. For each node \noderep, we aggregate the message from neighbouring node and edges by
\begin{align}
    \mathbf{m}_{i}^{t} &= \alpha \underbrace{\mathbf{W}_{nn} \left( \sum_{j \in \mathcal{N}_{i}} \mathbf{n}_{j}^{t-1} \right)}_\text{node to node message} + (1-\alpha) \underbrace{\mathbf{W}_{en} \left( \sum_{j \in \mathcal{N}_{i}} \mathbf{e}_{j \rightarrow i}^{t-1} \right)}_\text{edge to node message}, \label{eq:acc_attn}
\end{align}
where $\mathbf{W}_{nn}$ and $\mathbf{W}_{en}$ are the kernel matrices for node-to-node and node-to-edge communication and $0 \le \alpha \le 1$ is a hyper-parameter that controls the contribution of messages from edges and nodes.
Similarly, the message passing for edges are given by
\begin{align}
    \mathbf{d}_{i \rightarrow j}^{t} &= \mathbf{W}_{ee}[\mathbf{n}_{i}^{t-1} \mathbin\Vert \mathbf{n}_{j}^{t-1}],  \label{eq:edge_acc}
\end{align}
where $\mathbf{W}_{ee}$ is the kernel matrix for node-to-edge communication. Note that the message passing for edges is direction aware i.e. $\mathbf{d}_{i \rightarrow j} \neq \mathbf{d}_{j \rightarrow i}$. This is crucial as the relationship between two nodes change depending on direction of the edge for example \texttt{<cat, has, tail>} and \texttt{<tail, on, cat>}.
The incoming messages are combined with the states using a gating mechanism \cite{DBLP:journals/corr/LiTBZ15}.
\subsubsection{Pooling Layer}
We use gated pooling layers to generate vector representations of the two graphs. The pooling operation is given by
\begin{align}
    \mathbf{N} &= \sum_{k} f_{gate}(\mathbf{n}_{k}) \odot \mathbf{n}_{k} \\
    \mathbf{E} &= \sum_{ij} g_{gate}(\mathbf{e}_{i \rightarrow j}) \odot \mathbf{e}_{i \rightarrow j}
\end{align}
where $f_{gate}$ and $g_{gate}$ are gating functions that map node and edge states to a scalar and $\odot$ represents element-wise multiplication. These two vectors are then passed through a linear layer, after concatenation, to obtain the final vector representation of the graph. In the image graph \imagegraph, since there are no edge features, we use the pooled node vector $\mathbf{N}$ as the vector representation of the graph. 
\renewcommand{\arraystretch}{1.1}
\begin{table*}[t]
\centering
\resizebox{\textwidth}{!}{
\begin{tabular}{@{}lccccccccccc@{}}
\toprule
\toprule
                                         &                                &               & \multicolumn{3}{c}{Predicate Classification}     & \multicolumn{3}{c}{Scene Graph Classification}   & \multicolumn{3}{c}{Scene Graph Detection}     \\ \cmidrule(l){4-12} 
Dataset                                  & Model                          & Method        & mR@20          & mR@50          & mR@100         & mR@20          & mR@50          & mR@100         & mR@20         & mR@50         & mR@100        \\ \midrule
\multirow{8}{*}{Visual Genome}           & \multirow{2}{*}{VCTree \cite{tang2019learning}}        & Cross Entropy & 13.07          & 16.53          & 17.77          & 8.5            & 10.53          & 11.24          & 5.31          & 7.16          & 8.35          \\
                                         &                                & EBM-Loss      & \textbf{14.2}  & \textbf{18.19} & \textbf{19.72} & \textbf{10.4}  & \textbf{12.54} & \textbf{13.45} & \textbf{5.67} & \textbf{7.71} & \textbf{9.1}  \\ \cmidrule(l){2-12} 
                                         & \multirow{2}{*}{Motif \cite{zellers2018neural}}         & Cross Entropy & 12.45          & 15.71          & 16.8          & 6.95           & 8.85           & 9.05           & 5.07          & 6.91          & 8.12          \\
                                         &                                & EBM-Loss      & \textbf{14.17} & \textbf{18.02} & \textbf{19.53} & \textbf{8.18}  & \textbf{10.22} & \textbf{10.98} & \textbf{5.66} & \textbf{7.72} & \textbf{9.27} \\ \cmidrule(l){2-12} 
                                         & \multirow{2}{*}{IMP \cite{xu2017scene}}           & Cross Entropy & 8.85           & 10.97          & 11.77          & 5.4            & 6.4            & 6.74           & 2.2           & 3.29          & 4.14          \\
                                         &                                & EBM-Loss      & \textbf{9.43}  & \textbf{11.83} & \textbf{12.77} & \textbf{5.66}  & \textbf{6.81}  & \textbf{7.17}  & \textbf{2.78} & \textbf{4.23} & \textbf{5.44} \\ \cmidrule(l){2-12} 
                                         & \multirow{2}{*}{VCTree-TDE \cite{tang2020unbiased}} & Cross Entropy & 16.3           & 22.85          & 26.26          & 11.85          & 15.81          & 17.99          & 6.59          & 8.99          & 10.78         \\
                                         &                                & EBM-Loss      & \textbf{19.87} & \textbf{26.66} & \textbf{29.97} & \textbf{13.86} & \textbf{18.2}  & \textbf{20.45} & \textbf{7.1}  & \textbf{9.69} & \textbf{11.6} \\ \midrule
\multicolumn{1}{c}{\multirow{6}{*}{GQA}} & \multirow{2}{*}{Transformer \cite{vaswani2017attention}}   & Cross Entropy & 1.17           & 2.48           & 3.69           & .54            & .97            & 1.29           & -             & -             & -             \\
\multicolumn{1}{c}{}                     &                                & EBM-Loss      & \textbf{1.28}  & \textbf{2.94}  & \textbf{4.71}  & \textbf{.68}   & \textbf{1.32}  & \textbf{1.77}  & -             & -             & -             \\ \cmidrule(l){2-12} 
\multicolumn{1}{c}{}                     & \multirow{2}{*}{Motif \cite{zellers2018neural}}         & Cross Entropy & .85            & 1.8            & 2.75           & .42            & .81            & 1.18           & -             & -             & -             \\
\multicolumn{1}{c}{}                     &                                & EBM-Loss      & \textbf{.94}   & \textbf{2.1}   & \textbf{3.19}  & \textbf{.57}   & \textbf{.9}    & \textbf{1.26}  & -             & -             & -             \\ \cmidrule(l){2-12} 
\multicolumn{1}{c}{}                     & \multirow{2}{*}{IMP \cite{xu2017scene}}           & Cross Entropy & .5             & .94            & 1.32           & .28            & .5             & .65            & -             & -             & -             \\
\multicolumn{1}{c}{}                     &                                & EBM-Loss      & \textbf{.57}   & \textbf{1.07}  & \textbf{1.5}   & \textbf{.34}   & \textbf{.58}   & \textbf{.76}   & -             & -             & -             \\ \bottomrule \bottomrule
\end{tabular}
}
\caption{{\bf Quantitative Results.} We compare the proposed energy-based loss formulation against traditional cross-entropy loss using various state-of-the-art models. We report the mean Recall@K \cite{tang2019learning} under all three experimental setting.}
\label{tabel:quantative_results}
\vspace{-0.1in}
\end{table*}

\section{Experiments}
We present experimental results on two datasets: the Visual Genome dataset \cite{krishna2017visual} and the GQA dataset \cite{hudson2018gqa}.

\vspace{0.08in}
\noindent
\textbf{Visual Genome:} We use the pre-processed version of the dataset from \cite{xu2017scene}. The dataset consists of 108k images and contains 150 object categories and 50 predicate categories
We use the original split with $70\%$ of the images in the training set and the remaining $30\%$ in the test set, with $5$k images from the training set held out for validation \cite{zellers2018neural}.

\vspace{0.08in}
\noindent
\textbf{GQA:} The GQA dataset \cite{hudson2018gqa} is also constructed from images in the Visual Genome dataset. 
Starting from the scene graph annotations provided in Visual Genome, a normalization process is applied. This normalization process 
augments object and relation annotations and prunes inaccurate or unnatural relations. The resulting dataset contains a total of $1704$ object categories, $311$ relation categories 
We use the same $70-30$ split for the train and the test set, with $5$k images in the validation set. Compared to Visual Genome, the GQA dataset has denser graphs with a larger number of object and relation categories. 

\subsection{Scene Graph Generation Models}
The energy-based training introduced in this paper is generic and does not make any assumptions on the underlying scene graph generation model. This allows freedom in choosing the model architecture for image to scene graph mapping. On the Visual Genome dataset, we experiment with VCTree \cite{tang2019learning}, Neural Motifs \cite{zellers2018neural} and Iterative Message Passing \cite{xu2017scene}. We also experiment with VCTree-TDE \cite{tang2020unbiased}, where the inference involves counterfactual reasoning. On the GQA dataset, we experiment with Transformers \cite{vaswani2017attention}, instead of VCTree, as the larger number of object classes in the GQA dataset leads to considerably larger memory requirement in VCTree. The ability to experiment with different models demonstrates the versatility of our approach.

\subsection{Evaluation}

\noindent
\textbf{Relationship Recall (RR)}. We use the \textbf{mean Recall@K (mR@K)} metric \cite{tang2019learning} to evaluate the performance of the scene graph generation models. We report the mean Recall@K instead of the Regular Recall@K (R@K) due to data imbalance that leads to reporting bias as pointed out in recent works \cite{tang2020unbiased}. The evaluation is performed under three settings, (1) Predicate Classification (\textbf{PredCls}): Predict the relationship labels, given the image, object bounding boxes and object labels. (2) Scene Graph Classification (\textbf{SGCls}): Predict the object and predicate labels, given the image and bounding boxes and (3) Scene Graph Detection (\textbf{SGDet}): Predict the scene graph from the image.

\vspace{0.05in}
\noindent
\textbf{Zero-Shot Recall (zsR@K)}. Introduced in \cite{lu2016visual}, zsR@K evaluates the ability 
to identify subject-predictae-object relation triplets 
that were not observed during training. 
We compute zsR@K for 3 settings: PredCls, SGCls and SGDet.

\vspace{0.05in}
\noindent
\textbf{Few-Shot Recall}. We introduce the few-shot Recall@K (fsR@K) metric that reports the Recall@K for relation triplets that occur a certain number of times in the training set. Unlike the conventional few shot metric, we use a range of values to generate the few shot triplet splits. Thus instead of splitting the triplets into $1$-shot, $2$-shot, \etc, we split them into groups of $1-5$-shot, $6-10$-shot, \etc with triplets in a $k_1-k_2$-shot occurring between $k_1$ and $k_2$ times. 

\vspace{0.05in}
\noindent
\textbf{Sentence-to-Graph Retrieval (S2GR)}: Introduced in \cite{tang2020unbiased}, S2GR was designed to address the inability of RR (Relationship Recall) and zsR (Zero-Shot Recall) to capture graph level coherence. In S2GR, the scene graph predicted for an image (obtained in the SGDet setting) is used as a semantic representation of the image. The task then is to retrieve images using their graph representation, with the image caption as a query. Note that retrieval task is based solely on the detected scene graph and no other visual information. As a consequence, any bias in the scene graph generation will result in a decrease in the S2GR performance.  Similar to \cite{tang2020unbiased}, we report Recall@20/100 on 1k/5k gallery.

\subsection{Implementation Details}
\noindent
\textbf{Detector}. We pre-train a Faster R-CNN \cite{ren2015faster} with ResNeXt-101-FPN \cite{massa2018mrcnn, xie2017aggregated} backbone and freeze the weights of the network prior to scene graph generation training. 
We obtain the weights of the pre-trained detector for Visual Genome from \cite{tang2020sggcode}. For GQA, we pre-train the object detector using standard Faster-RCNN settings. The object detector has $28$ mAP on Visual Genome and $10$ mAP on GQA.

\vspace{0.06in}
\noindent
\textbf{Scene Graph Generator}. 
The baseline models, trained with standard cross entropy loss, as well as our proposed framework, are trained using an identical setup. We use an SGD optimizer with an initial learning rate of $10^{-2}$. For the Visual Genome models, we incorporate frequency bias \cite{zellers2018neural} into the training and inference.  We do not use the frequency bias on GQA dataset, due to  high memory requirements. 

\vspace{0.06in}
\noindent
\textbf{Energy Model}. In the sampling step of SGLD (Eq. \ref{eq:sgld}), we set the number of iterations ($\tau$) to $20$ and the number of message passing iterations in EGNN and GNN to $3$. We use a step size of $1$ and clip the gradients within $[-0.01, 0.01]$. After every gradient update step, we normalize the node and edge states of the scene graph to the range of $[0,1]$.

\vspace{0.06in}
\noindent
\textbf{Sentence-to-Graph Retrieval.} We use the same formulation for S2GR as previous work \cite{tang2020unbiased}. The problem is formulated as a graph matching problem where the image caption is converted to graph structure using \cite{krishna2017visual}. The scene graphs and text graphs are mapped into a joint embedding space for retrieval using a Bilinear Attention Network \cite{kim2018bilinear}. 


\renewcommand{\arraystretch}{1.1}
\begin{table}[t]
\resizebox{\linewidth}{!}{
\begin{tabular}{@{}cccccc@{}}
\toprule
\toprule
                               &                              &          & PredCls              & SGCls              & SGDet              \\ \cmidrule(l){4-6} 
Dataset                        & Model                        & Method   & zsR@20/50            & zsR@20/50          & zsR@20/50          \\ \midrule
\multirow{8}{*}{VG} & \multirow{2}{*}{VCTree}      & CE  & 1.43/4               & .39/1.2            & .19/.46            \\
                               &                              & EB & \textbf{2.25/5.36}   & \textbf{.87/1.87}  & \textbf{.21/.54}   \\ \cmidrule(l){2-6} 
                               & \multirow{2}{*}{Motif}       & CE  & 1.28/3.56            & .39/.83            & 0/.04              \\
                               &                              & EB & \textbf{2.07/4.87}   & \textbf{.52/1.25}  & \textbf{.11/.23}   \\ \cmidrule(l){2-6} 
                               & \multirow{2}{*}{IMP}         & CE  & 12.17/17.66          & 2.09/3.3           & .14/.39            \\
                               &                              & EB & \textbf{12.6/18.6}   & \textbf{2.29/3.7}  & \textbf{.16/.43}   \\ \cmidrule(l){2-6} 
                               & \multirow{2}{*}{VCTree-TDE}  & CE  & 8.98/14.52           & 3.16/4.97          & 1.47/2.3           \\
                               &                              & EB & \textbf{9.58/15.14}  & \textbf{4.18/6.38} & \textbf{1.62/2.68} \\ \midrule
\multirow{6}{*}{GQA}           & \multirow{2}{*}{Transformer} & CE  & 19.55/33.33          & .94/1.83           & -                  \\
                               &                              & EB & \textbf{20.11/34.33} & \textbf{1.2/2.05}  & -                  \\ \cmidrule(l){2-6} 
                               & \multirow{2}{*}{Motif}       & CE  & 17.74/30.61          & 1.27/2.16          & -                  \\
                               &                              & EB & \textbf{19.47/33.45} & \textbf{1.49/2.48} & -                  \\ \cmidrule(l){2-6} 
                               & \multirow{2}{*}{IMP}         & CE  & 15.58/27.6           & 1.02/1.88          & -                  \\
                               &                              & EB & \textbf{16.65/27.77} & \textbf{1.1/1.98}  & -                  \\ \bottomrule \bottomrule
\end{tabular}
}
\caption{\textbf{Zero-shot Recall.} The zero shot recall performance comparison of model trained using cross-entopy (CE) and energy-based loss (EB) on the Visual Genome (VG) and GQA dataset.}
\label{table:zeroshot}
\vspace{-0.1in}
\end{table}

\section{Experimental Results}

\noindent
{\bf Quantitative Results:}  
Table \ref{tabel:quantative_results} compares performance of various state-of-the-art methods trained using cross-entropy and our energy-based loss on two datasets, Visual Genome and GQA. 
We observe that training using the proposed energy loss leads to a consistent improvement in the the mean Recall in all three tasks, for all the models. For the VCTree model we obtain a relative improvement of $12.7 \%$, $22.3\%$ and $5.6\%$ on mR@$20$ for PredCls, SGCls and SGDet respectively when compared to cross entropy based training. We obtain relative improvements of similar magnitudes with the Motif, IMP and VCTree-TDE models. In addition, using our proposed method with the VCTree-TDE \cite{tang2020unbiased} model, we achieve the new state-of-the-art performance on the Visual Genome dataset. On the GQA dataset, we present results of three models, Transformer, Motif and IMP. Similar to the Visual Genome dataset, we observe a consistent improvement in the mean Recall metric under PredCls and SGCls, with each of the models. We omit experiments on SGDet task due to low mAP of the underlying object detector. The proposed method leads to a relative improvement of $8.5\%$ and $25.92\%$ in the mR@$20$ metric for the PredCls and SGCls task when using the Transformer model for scene graph generation. 
The absolute performance on the GQA dataset is lower, compared to Visual Genome dataset due to larger number of object and relationship classes.
Due to memory constraints, we omit the frequency prior on the GQA dataset, which was shown to be highly effective on the Visual Genome dataset \cite{zellers2018neural}. 

We provide additional results along with individual relation recall in the \textbf{supplementary material}. We observe that energy-based models obtain larger improvement in relations that have fewer training annotations, compared to their cross-entropy based counterparts.

\vspace{0.06in}
\noindent
{\bf Zero-Shot Recall:}
Table \ref{table:zeroshot} reports zero-shot recall (zsR@$20$ and zsR@$50$) for all models. Similar to mR@K, we note consistent improvement on the zero-shot recall. We attribute this behaviour to our energy-based structure aware framework, which facilitates learning of models that are capable of performing global 
scene graph reasoning.

\vspace{0.06in}
\noindent
{\bf Few-shot Recall:}
In Section \ref{sec:introduction}, we hypothesized that the energy based learning bakes an inductive bias into the learning, thereby allowing the models to learn from less amounts of data. To test this hypothesis, we measure the few-shot Recall@$20$ for the VCTree model. We train a scene graph detection model using the proposed energy-based model as well as the standard cross-entropy loss. The result, as shown in Table \ref{table:kshot}, demonstrate that our training framework is able to provide a significant boost in performance in few-shot scenarios, when less data is available. This shows that the energy formulation provides a data efficient method for learning scene graph generation.

\begin{table}[]
\centering
\begin{tabular}{@{}cccccc@{}}
\toprule
\toprule
                       & \multicolumn{5}{c}{Few-Shot Recall@20} \\ \midrule
$k_{1}-k_{2}$ shot & 1-5    & 6-10  & 11-15 & 16-20 & 20-25 \\ \midrule
C.E.                    & 16.9   & 24.41 & 27.73 & 31.52 & 32.31 \\
E.B.M.                & \textbf{18.55}  & \textbf{25.22} & \textbf{28.1}  & \textbf{32.05} & \textbf{32.57} \\
\bottomrule \bottomrule
\end{tabular}
\caption{\textbf{Few-shot Recall@20.} Table compares the few short recall performance of a VCTree \cite{tang2019learning} model trained using cross-entropy and energy-based loss. We observe larger improvement in performance when data is scarce.}
\label{table:kshot}
\vspace{-0.1in}
\end{table}
\begin{table}[t]
\resizebox{\linewidth}{!}{
\begin{tabular}{@{}cccccccc@{}}
\toprule
\toprule
                           \multicolumn{8}{c}{Sentence to Graph Retrieval}                                                        \\ \midrule
\multicolumn{2}{l}{Gallery Size} & \multicolumn{3}{c}{1000}                      & \multicolumn{3}{c}{5000}                        \\ \midrule
                          &      & R@20          & R@50          & R@100         & R@20          & R@50           & R@100          \\ \midrule
\multirow{2}{*}{VCTree}   & CE   & 14            & 28.4          & 44.6          & 4.1           & 8.9            & 14.98          \\
                          & EBM  & \textbf{17.2} & \textbf{32.5} & \textbf{48.6} & \textbf{5}    & \textbf{10.96} & \textbf{18.52} \\ \midrule
\multirow{2}{*}{Motif}    & CE   & 15.5          & 29.4          & 46.7          & 4.56          & 9.7            & 17             \\
                          & EBM  & \textbf{19.2} & \textbf{32}   & \textbf{49.2} & \textbf{5.22} & \textbf{10.96} & \textbf{18.64} \\ \bottomrule
                        \bottomrule
\end{tabular}
}
\caption{\textbf{Sentence to Graph Retrieval.} We compare the scene graph retrieval performance on gallery of $1000$ and $5000$ images.}
\label{table:sentencetograph}
\vspace{-0.1in}
\end{table}
\renewcommand{\arraystretch}{1.05}
\begin{table}[t]
\centering
\begin{tabular}{@{}cccc@{}}
\toprule
\toprule
                                    & \multicolumn{3}{c}{Predicate Classification} \\ \midrule
\multicolumn{1}{c|}{Ablation}       & mR@20         & mR@50        & mR@100        \\ \midrule
\multicolumn{1}{c|}{0-steps}        & 14.18         & 18.14        & 19.66         \\
\multicolumn{1}{c|}{20-steps}       & \emph{14.2}          & \emph{18.19}        & \emph{19.72}         \\
\multicolumn{1}{c|}{40-steps}       & 14.37         & 18.18        & 19.81         \\
\multicolumn{1}{c|}{60-steps}       & 14.41         & 18.23        & 19.79         \\
\multicolumn{1}{c|}{120-steps}      & 14.59         & 19.29        & 20            \\ \midrule
\multicolumn{1}{c|}{No-Image-Graph} & 14.19         & 18.05        & 19.55         \\ \bottomrule
\bottomrule
\end{tabular}
\caption{\textbf{Ablation.} We experiment with the number of optimization steps ($\tau$) needed to estimate the energy loss and the effect of excluding image information in the energy model. All numbers were obtained using a VCTree \cite{tang2019learning} model.}
\label{table:ablation}
\vspace{-0.15in}
\end{table}
\begin{figure*}[t]
    \centering
    \includegraphics[width=\linewidth]{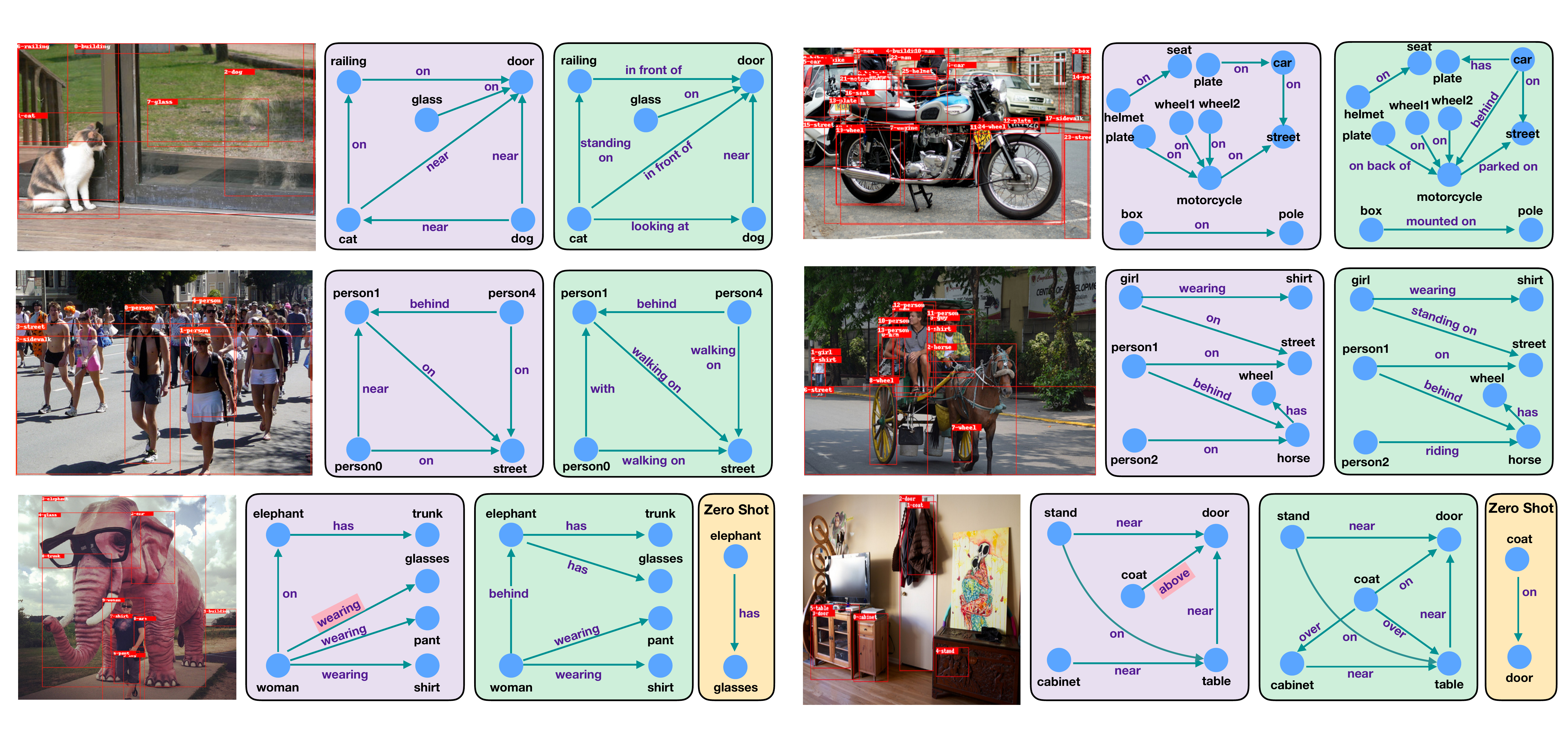}
    \vspace{-0.23in}
    \caption{\textbf{Qualitative Results.} Visualizations of scene graphs generated by a VCTree \cite{tang2019learning} model trained using cross-entropy loss (in purple) and proposed energy-based loss (in green). The top two rows show visualization of relation retrieval. The bottom row shows zero-shot relation retrieval results with the zero shot triplet show in yellow.}
    \label{fig:qulaitative_results}
    \vspace{-0.1in}
\end{figure*}

\vspace{0.06in}
\noindent
{\bf Sentence-to-Graph Retrieval:}
Table \ref{table:sentencetograph} compares the results of sentence-to-graph retrieval experiments. We use VCTree and Motif as our scene graph generation architecture. For each model we observe relative improvements ranging from 5\%-23\% in the retrieval recall compared to the scene graphs generated using corresponding baselines. 
This improvement in performance can be attributed to the more coherent and informative scene graphs generated by models trained using the proposed framework.

\vspace{0.06in}
\noindent
{\bf Ablation Studies:}
We investigated the impact of optimization in Eq.(\ref{eq:energy_loss}) on the effectiveness of the energy model. We experimented with a different number of optimization steps while training a predicate classification model using VCTree. Note that increasing the number of iterations corresponds to a more precise minima. We also investigated the effect of removing the image graph input to the energy model to determine the effectiveness of modeling the energy density over the joint space of scene and image graph. Table \ref{table:ablation} summarizes the mean recall@K for both of these experiments. We find that with an increase in the number of optimization steps in the energy loss, the mean recall almost consistently increases. Intuitively, this is expected as a larger number of optimization steps means that we have better chance of convergence when trying to find the minima in Eq.(\ref{eq:energy_loss}). This increase, however, comes with the added computation overhead and increase in training time. Similarly, we note that removing the image information from the joint modeling, there is a drop in performance as the model is now forced to learn from only the scene graph labels.

\vspace{0.06in}
\noindent
{\bf Qualitative Results:}
We visualize the qualitative results obtained from a VCTree model trained using the proposed energy-based framework as well as cross-entropy loss in Figure \ref{fig:qulaitative_results}. The top two rows show results from the regular relation retrieval task. We observe that the models trained using our proposed framework can consistently generate instructive relationships such as \texttt{mounted on}, \texttt{parked on}, \texttt{walking on}, \texttt{standing on} as opposed to the less informative and biased variant \texttt{on} generated by the baseline model. Similarly, in the top-left image, the energy-based training generates spatially informative relations such as \texttt{<cat, in front of, door>} instead of \texttt{<cat, near, door>} and \texttt{<cat, looking at, dog>} as opposed \texttt{<dog, near, cat>}. 

The bottom row shows results of zero-shot relation retrieval. In the first image, due to the triplets of \texttt{elephants} with \texttt{glasses} not being present in the training data, the baseline model predicts \texttt{<women, wearing, glasses>} whereas our method generates the accurate prediction \texttt{<elephant, has, glasses>}.

\section{Conclusion}
We present a novel model-agnostic energy-based learning framework for training scene graph generation models. Unlike cross-entropy based training, the proposed method embraces structure in the output space allowing the model to perform structure aware learning. We show that scene graph generation models can benefit from the proposed training framework by performing experiments on the Visual Genome and GQA datasets. We observe significant improvement in performance as compared to traditional cross-entropy based training. We also exhibit the generality and efficiency of our model through experiments in zero-shot and few-shot relationship settings. Finally, the proposed method does not make any assumptions on the underlying generation model and can be easily used with any model.

{\small
\bibliographystyle{ieee_fullname}
\bibliography{egbib}

\begin{thebibliography}{10}\itemsep=-1pt

\bibitem{du2019implicit}
Yilun Du and Igor Mordatch.
\newblock Implicit generation and modeling with energy based models.
\newblock In {\em Advances in Neural Information Processing Systems}, pages
  3608--3618, 2019.

\bibitem{grathwohl2020your}
Will Grathwohl, Kuan-Chieh Wang, Joern-Henrik Jacobsen, David Duvenaud,
  Mohammad Norouzi, and Kevin Swersky.
\newblock Your classifier is secretly an energy based model and you should
  treat it like one.
\newblock In {\em International Conference on Learning Representations}, 2020.

\bibitem{gu2019unpaired}
Jiuxiang Gu, Shafiq Joty, Jianfei Cai, Handong Zhao, Xu Yang, and Gang Wang.
\newblock Unpaired image captioning via scene graph alignments.
\newblock In {\em Proceedings of the IEEE/CVF Conference on Computer Vision and
  Pattern Recognition}, pages 10323--10332, 2019.

\bibitem{herzig2019learning}
Roei Herzig, Amir Bar, Huijuan Xu, Gal Chechik, Trevor Darrell, and Amir
  Globerson.
\newblock Learning canonical representations for scene graph to image
  generation.
\newblock In {\em Proceedings of the European Conference on Computer Vision
  (ECCV)}, 2020.

\bibitem{hudson2018gqa}
Drew~A Hudson and Christopher~D Manning.
\newblock Gqa: A new dataset for real-world visual reasoning and compositional
  question answering.
\newblock {\em Conference on Computer Vision and Pattern Recognition (CVPR)},
  2019.

\bibitem{johnson2018image}
Justin Johnson, Agrim Gupta, and Li Fei-Fei.
\newblock Image generation from scene graphs.
\newblock In {\em Proceedings of the IEEE/CVF Conference on Computer Vision and
  Pattern Recognition}, pages 1219--1228, 2018.

\bibitem{kim2018bilinear}
Jin-Hwa Kim, Jaehyun Jun, and Byoung-Tak Zhang.
\newblock Bilinear attention networks.
\newblock In {\em Advances in Neural Information Processing Systems}, pages
  1564--1574, 2018.

\bibitem{knyazev2020graph}
Boris Knyazev, Harm de Vries, C{\u{a}}t{\u{a}}lina Cangea, Graham~W Taylor,
  Aaron Courville, and Eugene Belilovsky.
\newblock Graph density-aware losses for novel compositions in scene graph
  generation.
\newblock {\em arXiv preprint arXiv:2005.08230}, 2020.

\bibitem{krishna2017visual}
Ranjay Krishna, Yuke Zhu, Oliver Groth, Justin Johnson, Kenji Hata, Joshua
  Kravitz, Stephanie Chen, Yannis Kalantidis, Li-Jia Li, David~A Shamma,
  Michael Bernstein, and Li Fei-Fei.
\newblock Visual genome: Connecting language and vision using crowdsourced
  dense image annotations.
\newblock {\em International journal of computer vision}, 123(1):32--73, 2017.

\bibitem{lecun2006tutorial}
Yann LeCun, Sumit Chopra, Raia Hadsell, M Ranzato, and F Huang.
\newblock A tutorial on energy-based learning.
\newblock {\em Predicting structured data}, 1(0), 2006.

\bibitem{lecun2005loss}
Yann LeCun and Fu~Jie Huang.
\newblock Loss functions for discriminative training of energy-based models.
\newblock In {\em AIStats}, volume~6, page~34. Citeseer, 2005.

\bibitem{li2018fnet}
Yikang Li, Wanli Ouyang, Zhou Bolei, Shi Jianping, Zhang Chao, and Xiaogang
  Wang.
\newblock Factorizable net: An efficient subgraph-based framework for scene
  graph generation.
\newblock In {\em Proceedings of the European Conference on Computer Vision
  (ECCV)}, 2018.

\bibitem{DBLP:journals/corr/LiTBZ15}
Yujia Li, Daniel Tarlow, Marc Brockschmidt, and Richard~S. Zemel.
\newblock Gated graph sequence neural networks.
\newblock In {\em 4th International Conference on Learning Representations,
  {ICLR}}, 2016.

\bibitem{Lin_2020_CVPR}
Xin Lin, Changxing Ding, Jinquan Zeng, and Dacheng Tao.
\newblock Gps-net: Graph property sensing network for scene graph generation.
\newblock In {\em Proceedings of the IEEE/CVF Conference on Computer Vision and
  Pattern Recognition (CVPR)}, June 2020.

\bibitem{lu2016visual}
Cewu Lu, Ranjay Krishna, Michael Bernstein, and Li Fei-Fei.
\newblock Visual relationship detection with language priors.
\newblock In {\em Proceedings of the European Conference on Computer Vision
  (ECCV)}, pages 852--869. Springer, 2016.

\bibitem{massa2018mrcnn}
Francisco Massa and Ross Girshick.
\newblock {maskrcnn-benchmark: Fast, modular reference implementation of
  Instance Segmentation and Object Detection algorithms in PyTorch}.
\newblock \url{https://github.com/facebookresearch/maskrcnn-benchmark}, 2018.

\bibitem{newell2017pixels}
Alejandro Newell and Jia Deng.
\newblock Pixels to graphs by associative embedding.
\newblock In {\em Advances in neural information processing systems}, pages
  2171--2180, 2017.

\bibitem{ren2015faster}
Shaoqing Ren, Kaiming He, Ross Girshick, and Jian Sun.
\newblock Faster r-cnn: Towards real-time object detection with region proposal
  networks.
\newblock In {\em Advances in neural information processing systems}, pages
  91--99, 2015.

\bibitem{song2019generative}
Yang Song and Stefano Ermon.
\newblock Generative modeling by estimating gradients of the data distribution.
\newblock In {\em Advances in Neural Information Processing Systems}, pages
  11918--11930, 2019.

\bibitem{tang2020sggcode}
Kaihua Tang.
\newblock A scene graph generation codebase in pytorch, 2020.
\newblock \url{https://github.com/KaihuaTang/Scene-Graph-Benchmark.pytorch}.

\bibitem{tang2020unbiased}
Kaihua Tang, Yulei Niu, Jianqiang Huang, Jiaxin Shi, and Hanwang Zhang.
\newblock Unbiased scene graph generation from biased training.
\newblock In {\em Proceedings of the IEEE/CVF Conference on Computer Vision and
  Pattern Recognition}, pages 3716--3725, 2020.

\bibitem{tang2019learning}
Kaihua Tang, Hanwang Zhang, Baoyuan Wu, Wenhan Luo, and Wei Liu.
\newblock Learning to compose dynamic tree structures for visual contexts.
\newblock In {\em Proceedings of the IEEE/CVF Conference on Computer Vision and
  Pattern Recognition}, pages 6619--6628, 2019.

\bibitem{vaswani2017attention}
Ashish Vaswani, Noam Shazeer, Niki Parmar, Jakob Uszkoreit, Llion Jones,
  Aidan~N Gomez, {\L}ukasz Kaiser, and Illia Polosukhin.
\newblock Attention is all you need.
\newblock In {\em Advances in neural information processing systems}, pages
  5998--6008, 2017.

\bibitem{welling2011bayesian}
Max Welling and Yee~W Teh.
\newblock Bayesian learning via stochastic gradient langevin dynamics.
\newblock In {\em Proceedings of the 28th international conference on machine
  learning (ICML-11)}, pages 681--688, 2011.

\bibitem{xie2017aggregated}
Saining Xie, Ross Girshick, Piotr Doll{\'a}r, Zhuowen Tu, and Kaiming He.
\newblock Aggregated residual transformations for deep neural networks.
\newblock In {\em Proceedings of the IEEE/CVF Conference on Computer Vision and
  Pattern Recognition}, pages 1492--1500, 2017.

\bibitem{xu2017scene}
Danfei Xu, Yuke Zhu, Christopher~B Choy, and Li Fei-Fei.
\newblock Scene graph generation by iterative message passing.
\newblock In {\em Proceedings of the IEEE/CVF Conference on Computer Vision and
  Pattern Recognition}, pages 5410--5419, 2017.

\bibitem{yang2018graph}
Jianwei Yang, Jiasen Lu, Stefan Lee, Dhruv Batra, and Devi Parikh.
\newblock Graph r-cnn for scene graph generation.
\newblock In {\em Proceedings of the European Conference on Computer Vision
  (ECCV)}, pages 670--685, 2018.

\bibitem{yang2019auto}
Xu Yang, Kaihua Tang, Hanwang Zhang, and Jianfei Cai.
\newblock Auto-encoding scene graphs for image captioning.
\newblock In {\em Proceedings of the IEEE/CVF Conference on Computer Vision and
  Pattern Recognition}, pages 10685--10694, 2019.

\bibitem{ZareianKC20}
Alireza Zareian, Svebor Karaman, and Shih{-}Fu Chang.
\newblock Bridging knowledge graphs to generate scene graphs.
\newblock In {\em Proceedings of the European Conference on Computer Vision
  (ECCV)}, pages 606--623. Springer, 2020.

\bibitem{ZareianWYC20}
Alireza Zareian, Zhecan Wang, Haoxuan You, and Shih{-}Fu Chang.
\newblock Learning visual commonsense for robust scene graph generation.
\newblock In {\em Proceedings of the European Conference on Computer Vision
  (ECCV)}, pages 642--657. Springer, 2020.

\bibitem{zellers2018neural}
Rowan Zellers, Mark Yatskar, Sam Thomson, and Yejin Choi.
\newblock Neural motifs: Scene graph parsing with global context.
\newblock In {\em Proceedings of the IEEE/CVF Conference on Computer Vision and
  Pattern Recognition}, pages 5831--5840, 2018.

\bibitem{zhang2019graphical}
Ji Zhang, Kevin~J Shih, Ahmed Elgammal, Andrew Tao, and Bryan Catanzaro.
\newblock Graphical co ntrastive losses for scene graph parsing.
\newblock In {\em Proceedings of the IEEE/CVF Conference on Computer Vision and
  Pattern Recognition}, pages 11535--11543, 2019.

\end{thebibliography}
}

\end{document}